# A novel hybrid model based on multi-objective Harris hawks optimization algorithm for daily PM$_{2.5}$ and PM$_{10}$ forecasting

Pei Du, Jianzhou Wang*, Yan Hao, Tong Niu, Wendong Yang

*School of Statistics, Dongbei University of Finance and Economics, Dalian 116025, China*

**Abstract:**

High levels of air pollution may seriously affect people's living environment and even endanger their lives. In order to reduce air pollution concentrations, and warn the public before the occurrence of hazardous air pollutants, it is urgent to design an accurate and reliable air pollutant forecasting model. However, most previous research have many deficiencies, such as ignoring the importance of predictive stability, and poor initial parameters and so on, which have significantly effect on the performance of air pollution prediction. Therefore, to address these issues, a novel hybrid model is proposed in this study. Specifically, a powerful data preprocessing techniques is applied to decompose the original time series into different modes from low- frequency to high-frequency. Next, a new multi-objective algorithm called MOHHO is first developed in this study, which are introduced to tune the parameters of ELM model with high forecasting accuracy and stability for air pollution series prediction, simultaneously. And the optimized ELM model is used to perform the time series prediction. Finally, a scientific and robust evaluation system including several error criteria, benchmark models, and several experiments using six air pollutant concentrations time series from three cities in China is designed to perform a compressive assessment for the presented hybrid forecasting model. Experimental results indicate that the proposed hybrid model can guarantee a more stable and higher predictive performance compared to others, whose superior prediction ability may help to develop effective plans for air pollutant emissions and prevent health problems caused by air pollution.

# 1 Introduction

In this section, the research background, literature review and the aims and innovation of this paper are shown in detail.

## *1.1 Background*

Atmospheric environmental pollution issues have become a significant challenge as well as received increasing attention by the public and policy makers around the world, particularly the developing countries, due to the thick smog and haze along with the dominant pollutant, particularly PM (particulate matter with aerodynamic diameter below 2.5 mm and below 10) frequently happened that contain many poisonous and harmful substances, which can cause all kinds of adverse health impacts such as respiratory diseases, heart failure, cardiovascular disease, etc. [1-3].

In recent years, smog and dust-haze caused by atmospheric pollution have gradually become the norm and covered most of China's areas including the Jing-Jin-Ji region, the Yangtze River delta region, the north of Southern China and other regions [4, 5]. Meanwhile, most major cities' concentrations of air pollutants greatly exceed standards recommended by the WHO (World Health Organization) and they have been regarded as the fourth biggest threat to the health of Chinese people after heart disease, dietary risk and smoking [6]. For instance, researches by the World Bank, WHO, and the Chinese Academy for Environmental Planning on the effect of air pollution on health summarized that between 350,000 and 500,000 people die prematurely every year on account of outdoor atmospheric pollution [7]. Moreover, the 2014 Environment Performance Index denoted that China ranked only 118th in environmental quality among the 178 observed countries [8]. Therefore, a series of relative measures, analysis, and precise prediction should be highly desirable to implement, which can not only supply some suggestions for the air quality regulatory requirements to reduce air pollution emission, but also help to guide people's daily activities, warn and protect them from harmful air pollutants.

## *1.2 Literature overview*

Air pollutant concentration prediction has become one of the most effective ways to assist relevant departments to develop effective management measures of air quality [9], which can supply the real-time data for forecasting concentration and help decision-makers analyze, monitoring air quality and warn people in time. Nowadays, many different prediction models have been designed and used to perform air pollutants forecasting. Generally, these models mainly contains three types: chemical transport models (CTMs), statistical methods and hybrid approaches [10].

Specifically speaking, owing to the high complexity of atmospheric chemical and transport processes, CTMs such as cannot always perform well. Different from CTMs, statistical methods are relatively simple and easy to operate, which have been widely applied to air pollutants prediction. For example, Zhang et al. applied ARIMA to analyze and predict Fuzhou's $PM_{2.5}$ [11]. Moreover, other statistical models such as grey models [12-14] are also studied in many literatures. However, these models are unable to accurately forecast due to their linear correlation structure. Hence, the nonlinear models are increasingly regarded as the alternatives of the linear ones [15]. The nonlinear models such as ANNs (artificial neural networks) [16], LSSVM (least squares support vector machine) [17] and other models are commonly used to air pollutants prediction because of their capacity of nonlinear mapping and self-learning.

However, the limited performance of the single models make it difficult to deal with several kinds of time series, particularly the datasets whose traits are unknown [18]. For instance, poor initial parameters in ANNs may result in poor prediction accuracy [19]. Under this background, the methods, called as hybrid models, which combine different techniques: forecasting models (ENN (Elman neural network), ELM (extreme learning machine), WNN (wavelet neutral network) etc.), optimization algorithms (HHO (Harris hawks optimization), GWO (Grey Wolf Optimizer), PSO (Particle Swarm Optimization) and so on) and data decomposition techniques (i.e. EMD (empirical mode decomposition), VMD (variational mode decomposition), CEEMD (complete ensemble empirical mode decomposition) etc.) [20]. Recently, hybrid prediction models have been popularly used to perform the air pollutants prediction, and improved the prediction performance. As an example, Niu et al. [5] developed a hybrid model through combining CEEMD and GWO, and support vector regression (SVR) to perform $PM_{2.5}$ concentration prediction. Wang et al. [21] combined the VMD, CEEMD, ELM tuned by DE (differential evolution) to predict air quality index in China. Additionally, Zhou et al. [22] also built a hybrid model applying data decomposition methods and GRNN (general regression neural network) to forecast $PM_{2.5}$ time series. Overall, the high performances of the hybrid models mentioned in the literatures showing that hybrid models possess better prediction abilities than that of single models.

In recent decades, most researchers have performed lots of studies concerning air pollution prediction and its related issues. Finally, these proposed models enhanced the forecasting effectiveness, which also provided the relative information of air pollution for decision-makers. However, there are still many deficiencies (such as using simple data preprocessing techniques, underestimating the importance of prediction stability, and sometimes adopting poor initial parameters, etc.) existing in the aforementioned models, which urgently need to be addressed. Hence, considering the drawbacks of previous studies about air pollutant forecasting aforementioned above, higher precise and more stable prediction models should be highly desirable to perform for air pollutant concentration prediction, which can not only supply some suggestions for the air quality regulatory requirements to reduce air pollution emission, but also help to guide people's daily activities, warn and protect them from harmful air pollutants.

*3. Aims and innovations*

In order to address the aforementioned shortcomings of previous studies and improve the prediction performance of air pollutant concentration forecasting, an innovative hybrid forecasting model through utilizing the improvement of accuracy and stability optimization strategy regarding ICEEMDAN (improved complete ensemble empirical mode decomposition with adaptive noise), MOHHO (multi-objective Harris hawks optimization) algorithm first presented here and ELM for air pollutant concentration prediction is performed in this study. Air pollutant concentrations i.e. $PM_{2.5}$ and $PM_{10}$ datasets collected from Jinan, Nanjing and Chongqing are taken as case studies to test the forecasting ability of the proposed hybrid model.

The leading innovations of this research are summarized as below: (i) The advantage of ICEEMDAN in decomposing air pollutant time series issues is employed to enhance air pollutant concentrations prediction accuracy; (ii) A novel algorithm, called MOHHO (multi-objective Harris hawks optimization), is successfully developed here, which outperforms MOGOA, MOPSO, and MSSA in most cases, it also supply a novel viable option for addressing multi-objective optimization issues; (iii) The innovative hybrid forecasting model based on ICEEMDAN, MOHHO and ELM are first developed, which can achieve high accurate and stable forecasting results simultaneously for air pollutant concentration prediction, it can also overcome the shortcomings of single objective optimization algorithms; (iv) A scientific and reasonable model assessment system including several experiments, seven model performance metrics, and DM (Diebold-Mariano) test applying several air pollution time series are utilized to give a systematic assessment of the presented hybrid model. Moreover, the superior performance of the proposed model indicates that the proposed hybrid model supplies a new option for the air quality regulatory requirements to reduce air pollution emission, but also help to guide people's daily activities, warn and protect them from harmful air pollutants.

### 1.4. The remainder structure of this paper

The structure of this research is designed as follows. Section 2 describes the framework of proposed model and the relative methodology formulation used in this study. Section 3 introduces the data description and model evaluation criteria adopted in this paper. And two experiments and forecasting results are built and analyzed in Section 4. Section 5 gives some discussions of the proposed model and the first developed MOHHO algorithm in this study. Finally, Section 6 concludes this study.

## 2 Methodology formulation

Subsection 2.1 describes the flowchart of the presented hybrid model, and the next several subsections i.e. subsection 2.2, 2.3 and 2.4 give the basic theories of the ICEEMDAN, MOHHO and ELM, respectively.

### *2.1 The framework of proposed model*

The flowchart of the hybrid forecasting model is displayed in **Fig. 1**.

- **Decomposition**. An effective data preprocessing techniques i.e. ICEEMDAN is adopted to decompose the original air pollutant concentration time series a finite set of modes shown in **Fig.1** (a). The different features hidden in the container throughput time series can be extracted by different modes from low- frequency to high- frequency.
- **Optimization**. This process contains (b), (c), (d) and (e) in **Fig. 1**. The (b) designs the input-output structure of the proposed model, the (c) is the flowchart of the MOHHO algorithm, which is used to optimize the ELM in (d), and then the optimized ELM can be built in (e). In this stage, the new presented MOHHO algorithm in this study is introduced to optimize the parameters of the ELM model with the hope of archiving high accuracy and stability for air pollutant concentrations prediction, simultaneously.
- **Prediction**. The ELM model optimized by the MOHHO are designed in (e) of **Fig.1** to perform the prediction of each modes obtained from ICEEMDAN. As a result, the final forecasts are achieved through integrating the forecasting results of each mode.
- **Evaluation**. As for the (f) in **Fig.1**, a scientific and robust evaluation system including several error criteria, benchmark models, and several experiments based on six air pollutant concentrations time series from China is designed to make a compressive evaluation for the proposed model with aim of deeply evaluating the forecasting ability of the presented hybrid forecasting model.

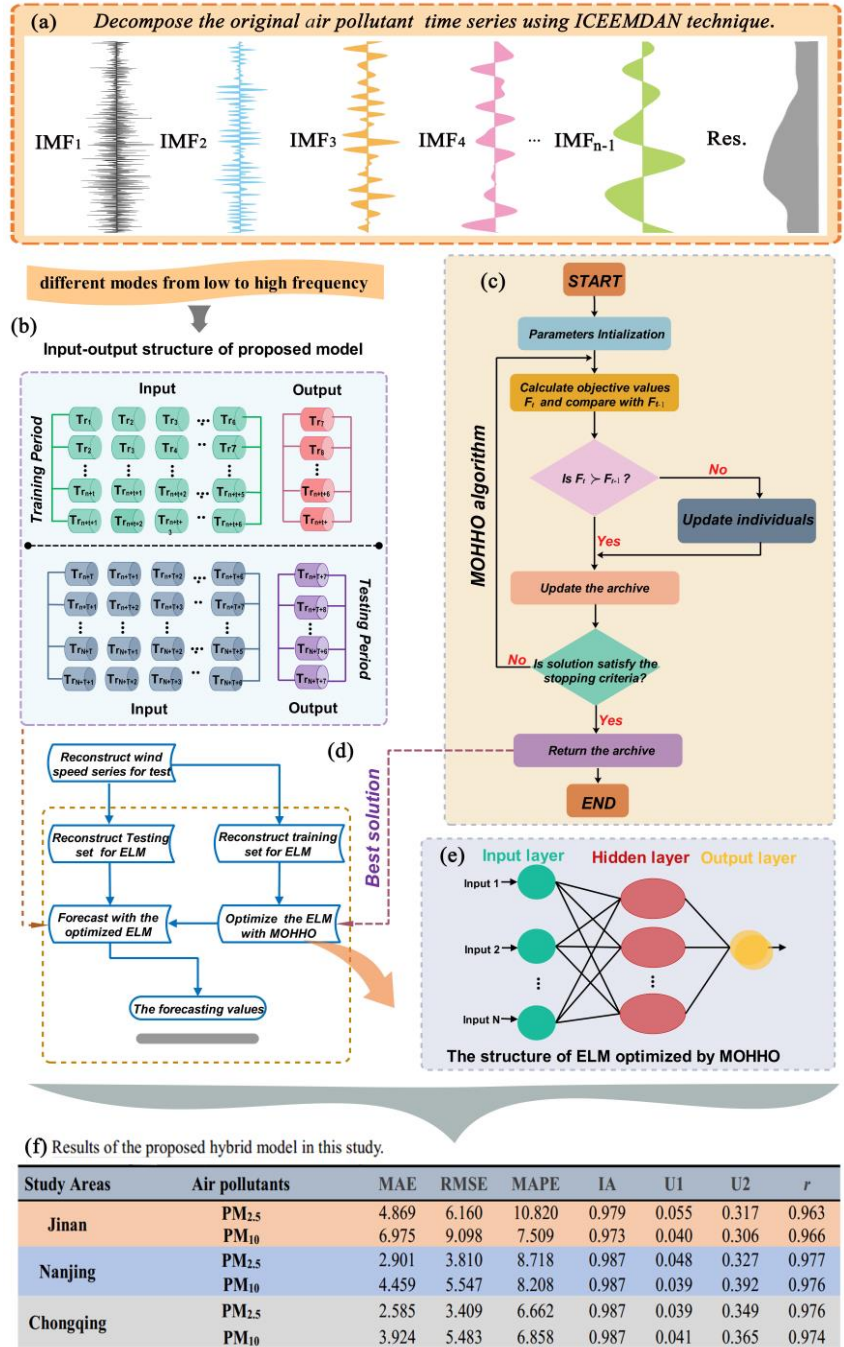

**Fig. 1.** The flowchart and forecasting results of the proposed hybrid model.

Moreover, the relative detailed algorithms and theories can be summarized primarily as below.

## *2.2 ICEEMDAN*

As one member of the EMD family, the ICEEMDAN in this study is used to decompose the original series of air pollutants, which is an improvement version developed by Colominas et al. [23]. Different form the conventional data processing methods, such as the wavelet decomposition restricted to nonstationary but linear data, Fourier decomposition primarily used to process smooth cyclical data, and so on, The EMD family is a series of data-processing approaches based on EMD, which are designed for nonlinear and nonstationary datasets [24].

Generally speaking, the EMD method proposed by Huang et al. [25] is an adaptive data driven technique that decomposes datasets into several intrinsic mode functions by a sifting process. Subsequently, noise-assisted versions are developed to address mode mixing problem of the EMD. Among them, EEMD method was proposed by Wu and Huang [26] to address the mode mixing issues through adding white noise. However, the addend white noise of EEMD cannot be completely removed. Therefore, another improvement version called CEEMD approach is developed, which can not only ensure the decomposition effect of EEMD, but also reduce the reconstruction error caused by adding paired noise with positive and negative signals [27]. However, there are still problems in CEEMD such as the completeness property is not proven, and the final averaging issue is still unsolved. As another improvement versions of EEMD, CEEMDAN [28] realizes a negligible reconstruction error and solve the modulus problem of different implementations of signal plus noise. Moreover, there are many aspects of CEEMDAN, which need to improve: (1) some residual noise, (2) spurious" modes in the early stages. Finally, the improved CEEMDAN is developed for achieving components with less noise and more physical meaning. This method have be widely used in many fields [29-32]. Details considering the ICEEMDAN technique refer to the relative literature [1].

### *2.3 Multi-objective Harris hawks optimization algorithm (MOHHO)*

The HHO (Harris hawks optimization) algorithm developed by Heidari et al. [33], was primarily inspired by the cooperative behavior and chasing style of Harris' hawks. Some hawks swooped cooperatively on their prey from different directions, trying to surprise it. Moreover, depend on the different scenes and patterns of prey flight, Hawks Harris can chose different chase patterns. There are three main phases in HHO: exploring a prey, surprise pounce, and many kinds of attacking strategies of Harris hawks. The details of each phase are descried as the follows.

### *Phase I: Exploration phase*

This subsection is mainly mathematically modeled to wait, search, and detect the prey. During every step, the Harris' hawks is regarded as the alternative or the best solutions. The *iter+1*-position $X(iter+1)$ of the Harris' hawks can be modeled based on the following equation:

$$X(iter+1) = \begin{cases} X_{rand}(iter) - r_1|X_{rand}(iter) - 2r_2 X(iter)| & if\ q \geq 0.5 \\ (X_{rabbit}(iter) - X_m(iter)) - r_3(LB + r_4(UB - LB)) & if\ q < 0.5 \end{cases} \quad (1)$$

Thereinto, $iter$ is the present iteration, the is $X_{rabbit}$ is the position of the rabbit, $X_{rand}$ is randomly chosen hawk at the present population, $r_i$, $i=1,2,3,4$, $q$ are random numbers between 0 and 1, and $X_m$ represents the average position of the hawks, which can be calculated by:

$$X_m(iter) = \frac{1}{N}\sum_{i=1}^{N} X_i(iter) \quad (2)$$

Where the vector $X_i$ represents the location of every hawk, and $N$ is the size of hawks.

### *Phase II: Transition from exploration to exploitation*

According to the escaping energy of the rabbit, HHO can be able to exchange exploration and exploitation. Moreover, the energy of the rabbit can be computed using the following formula:

$$E = 2E_0 \left(1 - \frac{iter}{T}\right) \tag{3}$$

In which the $E$ denotes the escaping energy of the rabbit, $T$ presents the maximum size about the iterations, and $E_0 \in (-1, 1)$ indicates the initial energy during every step. HHO can judge the state of rabbit depend on change trend of $E_0$ (The HHO start exploration phase in order to explore prey location, when $|E| \geq 1$, otherwise, this method attempt to exploit the neighborhood of the solutions during the exploitation steps).

*Phase III: Exploitation phase*

In this algorithm, hawks take a hard or soft besiege to hunt the prey from many directions softly or hard according to remaining energy of the prey. Under this situation, whether the prey escapes or not depend on the chance $r$ of a prey (successfully escaping: if $r < 0.5$). Moreover, if $|E| \geq 0.5$ HHO will adopt soft besiege, otherwise, the hard besiege will be applied. On the basis of the escaping phenomena of the prey and pursuing strategies of the Harris' hawks, HHO algorithm applied four strategies to simulate the attacking stage: soft besiege, hard besiege, soft besiege with progressive rapid dives, hard besiege with progressive rapid dives. Specially speaking, $|E| \geq 0.5$ is that the rabbit has enough energy to run away, however, whether the prey escape successfully or not depending on both values of the $|E|$ and $r$.

**(i) Soft besiege** ( $r \geq 0.5$ and $|E| \geq 0.5$ )

This measure can be written as follows:
$$X(iter+1) = \Delta X(iter) - E|JX_{rabbit}(iter) - X(iter)| \tag{4}$$
$$\Delta X(iter) = X_{rabbit}(iter) - X(iter) \tag{5}$$

Where $\Delta X(iter)$ indicates the difference between the position vector of the rabbit at *iter*- iteration, $J = 2(1-r_5)$ is the random jump intensity of the rabbit during the escaping process, and $r_5 \in (0,1)$ is a random number.

**(ii) Hard besiege** ( $r \geq 0.5$ and $|E| < 0.5$ )

As for this strategy, the present positions can be updated by the formula below:
$$X(iter+1) = \Delta X(iter) - E|JX_{rabbit}(iter) - X(iter)| \tag{6}$$

**(iii) Soft besiege with progressive rapid dives** ( $|E| \geq 0.5$ and $r < 0.5$ )

As for the soft besiege, the hawks decide their next step by the following equation:
$$Y = X_{rabbit}(iter) - E|JX_{rabbit}(iter) - X(iter)| \tag{7}$$

On the basis of the LF-based patterns, the hawks dive according to the following rules:
$$Z = Y + S \times LF(D) \tag{8}$$

In which $D$ donates the dimension of problem and $S_{1 \times D}$ indicates a random vector and the levy flight: *LF* can be calculated by:

$$LF(D)=0.01\times\frac{\mu\times\sigma}{|v|^{\frac{1}{\beta}}},\sigma=\left(\frac{\Gamma(1+\beta)\times\sin\left(\frac{\pi\beta}{2}\right)}{\Gamma\left(\frac{1+\beta}{2}\right)\times\beta\times 2^{\left(\frac{\beta-1}{2}\right)}}\right)^{\frac{1}{\beta}},\beta=1.5 \qquad (9)$$

Thereinto, $\mu$ and $v$ indicate random values between 0 and 1.

Therefore, the last strategy of this phase in order to update the hawks' positions can be expressed by:

$$X(iter+1)=\begin{cases} Y & if\ F(Y)<F(X(iter)) \\ Z & if\ F(Z)<F(X(iter)) \end{cases} \qquad (10)$$

**(iv) Hard besiege with progressive rapid dives** ($|E|<0.5$ and $r<0.5$)

In this step, the hawks are constantly close to the prey. The behavior can modeled as follows:

$$X(iter+1)=\begin{cases} Y & if\ F(Y)<F(X(iter)) \\ Z & if\ F(Z)<F(X(iter)) \end{cases} \qquad (11)$$

In which $Y$ and $Z$ can be computed using the following formulas:

$$Y=X_{rabbit}(iter)-E|JX_{rabbit}(iter)-X_{m}(iter)| \qquad (12)$$

$$Z=Y+S\times LF(D) \qquad (13)$$

where $X_{m}(iter)=\frac{1}{N}\sum_{i=1}^{N}X_{i}(iter)$.

In this subsection, a multi-objective version of the HHO, namely MOHHO (Multi-objective Harris hawks optimization algorithm), is developed for solving multi-objective optimization problem using HHO. In MOHHO algorithm, an archive as well as roulette wheel selection is utilized to carry out the multi-objective function. The archive is primary in charge of storing the non-dominated Pareto optimal solutions selected by the present iteration. What is noteworthy is that the maximum number of the archive need to set in advance, which represents the maximum capacity about the best non-dominated solutions obtained by fat. During each iteration, the obtained non-dominated solutions are used to compare with the archived members to update the archive. If a novel solution is better or at least no worse than the solutions in the archive, at last they will be recorded in the archive. More importantly, when the archive is full, a removed probability is also considered to delete the most populated neighborhoods for accommodating new solutions.

$$P_i=\frac{N_i}{c},\ c>1 \qquad (14)$$

In which $c$ represents a constant and $N_i$ is the number of solutions in the vicinity of the *i-th* solution.

Additionally, a roulette wheel method with probability $P_i=c/N_i$ is also introduced in MOHHO to improve the distribution of solutions in the archive. And the related concepts of multi-objective optimization are given in **Appendix A.**

*2.4 Extreme Learning Machine*

The ELM model, first developed by Huang et al [34], becomes famous and widely applied in forecasting field due to it has fast learning speed and its input weights and hidden biases are initialized with random numbers. Additionally, the output weights can be obtained using an inverse operation to the hidden layer output matrix [35]. More importantly, the ELM model is easily utilized to address least squares problems with much less time [36]. Considering $N$ observations: $(\boldsymbol{x}_t, \boldsymbol{y}_t)$, $t = 1, 2, .., N$, where $\boldsymbol{x}_t \in R^n$ represents input pattern, $\boldsymbol{y}_t \in R^m$ is the output. Then, the ELM's output with $L$ hidden neurons is expressed using the following formula:

$$\sum_{j=1}^{L} \boldsymbol{\beta}_j f\left(\boldsymbol{w}_j \boldsymbol{x}_t + b_j\right) = \boldsymbol{o}_j, \quad t = 1, 2, ..., N \tag{15}$$

Where $b_j$ is the bias, $f(\cdot)$ is the activation function, $\boldsymbol{w}_j = [w_{j1}, w_{j2}, ..., w_{jn}]^T$ and $\boldsymbol{\beta}_j = [\beta_{j1}, \beta_{j2}, ..., \beta_{jm}]^T$ represent the input weight and output weight vectors, respectively.

The Eq. (13) can also be written as below:

$$H\beta = T \tag{16}$$

where $H$ is the hidden layer output matrix, which can be expressed as below:

$$H = \begin{bmatrix} f(\boldsymbol{w}_1 \cdot \boldsymbol{x}_1 + b_1) & \cdots & f(\boldsymbol{w}_L \cdot \boldsymbol{x}_1 + b_L) \\ \vdots & \cdots & \vdots \\ f(\boldsymbol{w}_1 \cdot \boldsymbol{x}_N + b_1) & \cdots & f(\boldsymbol{w}_L \cdot \boldsymbol{x}_N + b_L) \end{bmatrix}_{N \times L} \tag{17}$$

$$\beta = \begin{bmatrix} \boldsymbol{\beta}_1^T \\ \vdots \\ \boldsymbol{\beta}_L^T \end{bmatrix}_{L \times m}, \quad T = \begin{bmatrix} t_1^T \\ \vdots \\ t_N^T \end{bmatrix}_{N \times m} \tag{18}$$

Moreover, the coefficient $\beta$ can be computed by following equation.

$$\left\| H\hat{\beta} - T \right\| = \left\| HH^+T - T \right\| = \min_{\beta} \left\| H\beta - T \right\| \tag{19}$$

In which $\hat{\beta} = H^+T$, the $H^+$ represents the Moore-Penrose generalized inverse of the hidden layer output matrix.

## 3 Data description and model evaluation criteria

This section contains of two parts: data description and model evaluation criteria, the details are given as follows.

### 3.1 Study area and data description

As for this subsection, the air quality datasets including daily $PM_{2.5}$ and $PM_{10}$ time series collected from three cities i.e. Jinan, Nanjing and Chongqing are taken as case studies to test the forecasting ability of the proposed hybrid model. Six historical daily air pollutant concentrations time series, whose descriptive statistics shown in **Table 1**, are all collected from November 1, 2017 to November 23, 2018 with a total 338 observations. The former 338 samples are taken as the training datasets to build the proposed model, and the remaining data series are applied to verify the prediction abilities of the forecasting models. Moreover, three study areas and their related information are shown in **Fig. 2**.

**Table 1.** Descriptive statistics of air pollutants (unit of $PM_{2.5}$ $PM_{10}$: $\mu g/m^3$).

| Areas | Pollutants | Maximum | Median | Mean | Minimum | Std. |
|---|---|---|---|---|---|---|
| **Jinan** | $PM_{2.5}$ | 269 | 44 | 8 | 52.77 | 35.33 |
| | $PM_{10}$ | 410 | 102.5 | 12 | 112.08 | 53.88 |
| **Nanjing** | $PM_{2.5}$ | 221 | 35 | 7 | 43.56 | 29.58 |
| | $PM_{10}$ | 287 | 67 | 11 | 78.34 | 42.32 |
| **Chongqing** | $PM_{2.5}$ | 156 | 34 | 11 | 41.66 | 26.02 |
| | $PM_{10}$ | 229 | 58 | 16 | 67.42 | 36.77 |

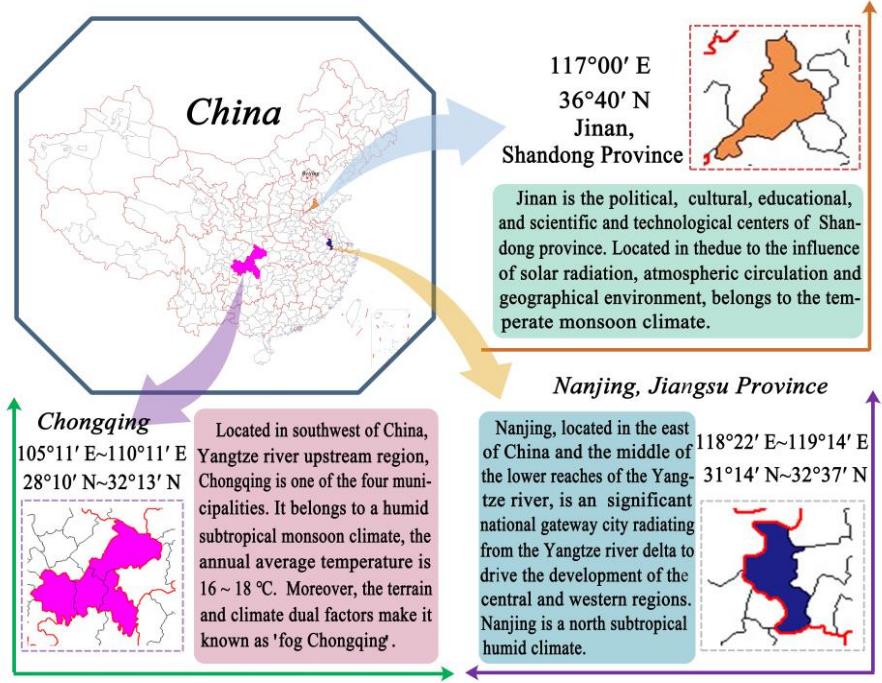

**Fig. 2** Information of the study areas in this study

*3.2 Model evaluation criteria*

Recently, many model evaluation criteria have been widely employed to verify the prediction performance of prediction models in many literatures. In this study, five common error criteria including the MAE (mean absolute error), RMSE (root mean square error), MAPE (mean absolute percent error), U1 (Theil U statistic 1), and U2 (Theil U statistic 2) [37], are adopted to evaluate the effective of the developed model and other models used for comparisons. The smaller the values of the error criteria are, the better the model is. Additionally, the IA (index of agreement) and *r* (Pearson's correlation coefficient) are also used in this study, the higher the values of IA and *r* are, the better the models are.

$$MAE = \frac{1}{L}\sum_{l=1}^{L}\left|y_{ACT(l)} - y_{ACT(l)}\right| \qquad (20)$$

$$RMSE = \sqrt{\frac{1}{L}\sum_{l=1}^{L}\left(y_{PRE(l)} - y_{ACT(l)}\right)^2} \qquad (21)$$

$$MAPE = \frac{1}{L}\sum_{l=1}^{L}\left|\frac{y_{PRE(l)} - y_{ACT(l)}}{y_{ACT(l)}}\right| \times 100\% \qquad (22)$$

$$IA = 1 - \frac{\sum_{l=1}^{L}\left(y_{PRE(l)} - y_{ACT(l)}\right)^2}{\sum_{l=1}^{L}\left(\left|y_{PRE(l)} - \overline{y}_{ACT}\right| + \left|y_{PRE(l)} - \overline{y}_{ACT}\right|^2\right)} \quad (23)$$

$$U1 = \frac{\sqrt{\frac{1}{L}\sum_{l=1}^{L}\left(y_{PRE(l)} - y_{ACT(l)}\right)^2}}{\sqrt{\frac{1}{L}\sum_{l=1}^{L}y_{ACT(l)}^2} + \sqrt{\frac{1}{L}\sum_{l=1}^{L}y_{PRE(l)}^2}} \quad (24)$$

$$U2 = \frac{\sqrt{\frac{1}{L}\sum_{l=1}^{L}\left(\frac{y_{PRE(l+1)} - y_{ACT(l+1)}}{y_{ACT(l)}}\right)^2}}{\sqrt{\frac{1}{L}\sum_{l=1}^{L}\left(\frac{y_{PRE(l)} - y_{ACT(l)}}{y_{ACT(l)}}\right)^2}} \quad (25)$$

$$r = \frac{\sum_{l=1}^{L}\left(y_{ACT(l)} - \overline{y}_{ACT}\right)\left(y_{PRE(l)} - \overline{y}_{PRE}\right)}{\sqrt{\sum_{l=1}^{L}\left(y_{ACT(l)} - \overline{y}_{ACT}\right)^2 \sum_{l=1}^{L}\left(y_{PRE(l)} - \overline{y}_{PRE}\right)^2}} \quad (26)$$

In which $L$ is the number of testing time series, $\overline{y}$ is the mean value of $y$, $y_{ACT(l)}$ is the $l$-th actual values, and $y_{PRE(l)}$ represents the $l$-th predicted values.

## 4 Experiments and analysis

Aiming to verify the validity of the developed hybrid model in this study, several datasets of air pollutant concentration and five comparison models are collected and established, respectively in this study. Furthermore, two experiments i.e. ***Experiment I: PM$_{2.5}$ forecasting***, and ***Experiment II: PM$_{10}$ forecasting***, are designed to further illustrate the superiority of the presented hybrid models. In addition, the improvement percentages of model evaluation criteria (listed in **Appendix B**) are also adopted in this section in order to discuss the performances of the prediction models. The details can be seen as follows.

### *4.1. Experiment I: PM$_{2.5}$ forecasting*

In order to show the superiority of the presented ICEEMDAN-MOHHO-ELM model, three PM$_{2.5}$ series datasets collected form Jinan, Nanjing and Chongqing are adopted in this experiment. Moreover, the ARIMA, ELM, LSSVM, EMD-MOHHO-ELM, and CEEMD-MOHHO-ELM models are built to compare with the proposed model. Moreover, this subsection design three types of model comparisons: comparisons between individual models, comparisons between hybrid models, and comparisons between individual models and hybrid models. The forecasting errors of the proposed model and comparison models are listed in **Table 2**, where the values marked in boldface are the best values of each evaluation metric.

With regard to the single models from **Table 2**, several conclusions can be drawn that the predictive errors of ARIMA, ELM, and LSSVM are much high, whose values of the MAPE are all bigger than 20%, they are difficult to satisfy the air quality regulatory requirements. For instance, as for Jinan, the MAPE of ARIMA model is 35.913%, the MAPE of ELM model is 38.441%, and the MAPE of LSSVM model is 38.945%, whereas the MAE is 15.866, 15.619 and 15.473. This phenomena show the limited performance of the single models once again.

As for hybrid model shown in **Table 2**, several conclusions can be obtained that different hybrid models possess different prediction accuracy, whose prediction performance maybe vary widely. For example, as for Nanjing, the RMSE values of EMD-MOHHO-ELM, CEEMD-MOHHO-ELM and ICEEMDAN-MOHHO-ELM are 23.567%, 13.112% and 8.718%, respectively. The reason can be interpreted by that the ICEEMDAN contribute more on the performance of the proposed hybrid model than the comparisons hybrid models with EMD and CEEMD. This aforementioned discussion show the necessity and importance of usage about the data decomposition technique.

The error values of single models and the hybrid models in **Table 2** indicating that hybrid models perform better than single models in most cases. As an example, with respect to Chongqing, the IA values of individual models are 0.812 (ARIMA), 0.803 (ELM) and 0.802 (LSSVM), while the IA values of hybrid models are 0.860 (EMD-MOHHO-ELM), 0.965 (CEEMD-MOHHO-ELM) and 0.986 (ICEEMDAN-MOHHO-ELM), respectively. The Comparative analysis between the presented hybrid model and the single model confirm the advantages of the hybrid forecasting model.

Moreover, **Table 3** displays the improvement percentages of PM$_{2.5}$ prediction among the developed hybrid model and the comparison models. According to **Tables 2** and **3**, it is obvious that the values of MAE, RMSE, MAPE, U1, and U2 of the developed hybrid model are all smaller than the other considered models and the values of IA and *r* of the developed hybrid model are all bigger than that of the comparison models, which further confirm the superiority of the presented hybrid model in terms of the prediction ability.

**Table 2**. The forecasting errors of different models for PM$_{2.5}$ in three cities.

| Areas | Models | MAE | RMSE | MAPE | IA | U1 | U2 | *r* |
|---|---|---|---|---|---|---|---|---|
| **Jinan** | **ARIMA** | 15.866 | 20.901 | 35.913 | 0.693 | 0.197 | 0.862 | 0.496 |
| | **ELM** | 15.619 | 20.251 | 38.441 | 0.699 | 0.184 | 0.682 | 0.494 |
| | **LSSVM** | 15.473 | 19.745 | 38.945 | 0.682 | 0.180 | 0.704 | 0.497 |
| | **EMD-MOHHO-ELM** | 9.947 | 11.257 | 24.447 | 0.920 | 0.101 | 0.658 | 0.865 |
| | **CEEMD-MOHHO-ELM** | 6.969 | 8.762 | 15.253 | 0.950 | 0.080 | 0.530 | 0.936 |
| | **ICEEMDAN-MOHHO-ELM** | **4.869** | **6.160** | **10.820** | **0.979** | **0.055** | **0.317** | **0.963** |
| **Nanjing** | **ARIMA** | 10.896 | 14.272 | 40.799 | 0.757 | 0.186 | 0.624 | 0.609 |
| | **ELM** | 10.541 | 13.841 | 37.986 | 0.765 | 0.175 | 0.586 | 0.624 |
| | **LSSVM** | 10.814 | 14.121 | 39.916 | 0.745 | 0.179 | 0.598 | 0.602 |
| | **EMD-MOHHO-ELM** | 6.627 | 7.905 | 23.567 | 0.947 | 0.099 | 0.593 | 0.900 |
| | **CEEMD-MOHHO-ELM** | 4.287 | 6.771 | 13.112 | 0.951 | 0.087 | 0.476 | 0.940 |
| | **ICEEMDAN-MOHHO-ELM** | **2.901** | **3.810** | **8.718** | **0.987** | **0.048** | **0.327** | **0.977** |
| **Chongqing** | **ARIMA** | 8.841 | 11.929 | 23.708 | 0.812 | 0.139 | 0.600 | 0.676 |
| | **ELM** | 9.157 | 12.069 | 24.922 | 0.803 | 0.139 | 0.617 | 0.665 |
| | **LSSVM** | 8.828 | 11.972 | 23.505 | 0.802 | 0.138 | 0.586 | 0.663 |
| | **EMD-MOHHO-ELM** | 7.922 | 10.128 | 22.450 | 0.860 | 0.115 | 0.753 | 0.764 |
| | **CEEMD-MOHHO-ELM** | 3.972 | 5.464 | 11.178 | 0.965 | 0.062 | 0.622 | 0.940 |
| | **ICEEMDAN-MOHHO-ELM** | **2.814** | **3.572** | **7.381** | **0.986** | **0.041** | **0.320** | **0.973** |

**Table 3.** Improvement percentages of the developed model compared with other models.

| Areas | Models | $P_{MAE}$ | $P_{RMSE}$ | $P_{MAPE}$ | $P_{IA}$ | $P_{U1}$ | $P_{U2}$ | $P_r$ |
|---|---|---|---|---|---|---|---|---|
| Jinan | Proposed model vs. ARIMA | 69.312 | 70.528 | 69.872 | 41.270 | 72.081 | 63.225 | 94.153 |
| | Proposed model vs. ELM | 68.826 | 69.582 | 71.853 | 40.057 | 70.109 | 53.519 | 94.939 |
| | Proposed model vs. LSSVM | 68.532 | 68.802 | 72.217 | 43.548 | 69.444 | 54.972 | 93.763 |
| | Proposed model vs. EMD-MOHHO-ELM | 51.051 | 45.278 | 55.741 | 6.413 | 45.545 | 51.824 | 11.329 |
| | Proposed model vs. CEEMD-MOHHO-ELM | 30.133 | 29.696 | 29.063 | 3.053 | 31.250 | 40.189 | 2.885 |
| Nanjing | Proposed model vs. ARIMA | 73.376 | 73.304 | 78.632 | 30.383 | 74.194 | 47.596 | 60.427 |
| | Proposed model vs. ELM | 72.479 | 72.473 | 77.049 | 29.020 | 72.571 | 44.198 | 56.571 |
| | Proposed model vs. LSSVM | 73.174 | 73.019 | 78.159 | 32.483 | 73.184 | 45.318 | 62.292 |
| | Proposed model vs. EMD-MOHHO-ELM | 56.225 | 51.803 | 63.008 | 4.224 | 51.515 | 44.857 | 8.556 |
| | Proposed model vs. CEEMD-MOHHO-ELM | 32.330 | 43.731 | 33.511 | 3.785 | 44.828 | 31.303 | 3.936 |
| Chongqing | Proposed model vs. ARIMA | 68.171 | 70.056 | 68.867 | 21.429 | 70.504 | 46.667 | 43.935 |
| | Proposed model vs. ELM | 69.269 | 70.404 | 70.384 | 22.790 | 70.504 | 48.136 | 46.316 |
| | Proposed model vs. LSSVM | 68.124 | 70.164 | 68.598 | 22.943 | 70.290 | 45.392 | 46.757 |
| | Proposed model vs. EMD-MOHHO-ELM | 64.479 | 64.731 | 67.122 | 14.651 | 64.348 | 57.503 | 27.356 |
| | Proposed model vs. CEEMD-MOHHO-ELM | 29.154 | 34.627 | 33.969 | 2.176 | 33.871 | 48.553 | 3.511 |

### *4.2. Experiment II: PM$_{10}$ forecasting*

In this subsection, three daily PM$_{10}$ time series of Jinan, Nanjing and Chongqing are used as another experiment to further verify the forecasting performance of the developed ICEEMDAN-MOHHO-ELM model. Similar to the prediction of PM$_{2.5}$, the five benchmark models are also taken as the comparison models. **Table 4** shows the values of the model evaluation criteria including MAE, RMSE, MAPE, IA, U1, U2 and *r* of the proposed model and comparison models adopted in this study. In addition, **Table 5** gives the improvement percentages of PM$_{10}$ prediction among the proposed model and comparison models.

It can be seen from **Tables 4** and **5** that the presented ICEEMDAN-MOHHO-ELM model is superior to the comparison models. Specifically, the developed model can obtain the smallest values of MAE, RMSE, MAPE, U1 and U2 and achieve largest values of IA and *r* when it compared with the other models. For example, as for Nanjing, the U1 values of are 0.143 (ARIMA), 0.140 (ELM), 0.139 (LSSVM), 0.073 (EMD-MOHHO-ELM), 0.125 (CEEMD-MOHHO-ELM) and 0.039 (CEEMD-MOHHO-ELM), respectively.

Based on the discussions and the values in **Tables 4** and **5**, it can be observed that as for the three daily PM$_{10}$ time series of Jinan, Nanjing and Chongqing, the proposed ICEEMDAN-MOHHO-ELM model can obtain the best values of MAE, RMSE, MAPE, IA, U1, U2 and r, showing that the advantages of the hybrid model and the importance usage of the data decomposition technique, especially more efficient decomposition method such as the ICEEMDAN. Taking the prediction results of series PM$_{10}$ in Jinan as an example, in **Table 5** the values of $P_{MAE}$ are 71.140% (Proposed model vs. ARIMA), 70.561% (Proposed model vs. ELM), 69.865% (Proposed model vs. LSSVM), 56.604% (Proposed model vs. EMD-MOHHO-ELM), 43.741% (Proposed model vs. CEEMD-MOHHO-ELM), respectively. This phenomenon illustrates that the developed ICEEMDAN-MOHHO-ELM model significantly outperforms the prediction accuracy compared to the five models.

**Table 4.** The forecasting errors of different models for $PM_{10}$ in three cities.

| Areas | Models | MAE | RMSE | MAPE | IA | U1 | U2 | r |
|---|---|---|---|---|---|---|---|---|
| Jinan | ARIMA | 24.168 | 31.721 | 25.061 | 0.592 | 0.143 | 0.839 | 0.347 |
| | ELM | 23.693 | 31.038 | 26.880 | 0.564 | 0.134 | 0.711 | 0.301 |
| | LSSVM | 23.146 | 30.402 | 26.236 | 0.545 | 0.133 | 0.720 | 0.300 |
| | EMD-MOHHO-ELM | 16.073 | 19.400 | 17.128 | 0.862 | 0.084 | 0.649 | 0.772 |
| | CEEMD-MOHHO-ELM | 12.398 | 15.559 | 13.339 | 0.907 | 0.068 | 0.606 | 0.874 |
| | ICEEMDAN-MOHHO-ELM | **6.975** | **9.098** | **7.509** | **0.973** | **0.040** | **0.306** | **0.966** |
| Nanjing | ARIMA | 15.957 | 19.760 | 31.894 | 0.767 | 0.143 | 0.862 | 0.635 |
| | ELM | 16.322 | 19.935 | 32.341 | 0.784 | 0.140 | 0.704 | 0.638 |
| | LSSVM | 16.077 | 19.830 | 32.707 | 0.771 | 0.139 | 0.726 | 0.633 |
| | EMD-MOHHO-ELM | 8.886 | 10.352 | 17.419 | 0.953 | 0.073 | 0.598 | 0.912 |
| | CEEMD-MOHHO-ELM | 9.304 | 17.243 | 14.872 | 0.856 | 0.125 | 0.770 | 0.746 |
| | ICEEMDAN-MOHHO-ELM | **4.459** | **5.547** | **8.208** | **0.987** | **0.039** | **0.392** | **0.976** |
| Chongqing | ARIMA | 14.066 | 18.365 | 25.362 | 0.810 | 0.140 | 0.585 | 0.681 |
| | ELM | 14.373 | 19.397 | 27.348 | 0.778 | 0.145 | 0.616 | 0.628 |
| | LSSVM | 14.031 | 18.760 | 25.789 | 0.795 | 0.140 | 0.547 | 0.655 |
| | EMD-MOHHO-ELM | 11.945 | 15.598 | 22.879 | 0.874 | 0.114 | 0.718 | 0.781 |
| | CEEMD-MOHHO-ELM | 6.067 | 7.587 | 10.629 | 0.973 | 0.056 | 0.504 | 0.951 |
| | ICEEMDAN-MOHHO-ELM | **3.924** | **5.483** | **6.858** | **0.987** | **0.041** | **0.365** | **0.974** |

**Table 5.** Improvement percentages of the developed model compared with other models.

| Areas | Models | $P_{MAE}$ | $P_{RMSE}$ | $P_{MAPE}$ | $P_{IA}$ | $P_{U1}$ | $P_{U2}$ | $P_r$ |
|---|---|---|---|---|---|---|---|---|
| Jinan | Proposed model vs. ARIMA | 71.140 | 71.319 | 70.037 | 64.358 | 72.028 | 63.528 | 178.386 |
| | Proposed model vs. ELM | 70.561 | 70.688 | 72.065 | 72.518 | 70.149 | 56.962 | 220.930 |
| | Proposed model vs. LSSVM | 69.865 | 70.074 | 71.379 | 78.532 | 69.925 | 57.500 | 222.000 |
| | Proposed model vs. EMD-MOHHO-ELM | 56.604 | 53.103 | 56.160 | 12.877 | 52.381 | 52.851 | 25.130 |
| | Proposed model vs. CEEMD-MOHHO-ELM | 43.741 | 41.526 | 43.706 | 7.277 | 41.176 | 49.505 | 10.526 |
| Nanjing | Proposed model vs. ARIMA | 72.056 | 71.928 | 74.265 | 28.683 | 72.727 | 54.524 | 53.701 |
| | Proposed model vs. ELM | 72.681 | 72.175 | 74.620 | 25.893 | 72.143 | 44.318 | 52.978 |
| | Proposed model vs. LSSVM | 72.265 | 72.027 | 74.904 | 28.016 | 71.942 | 46.006 | 54.186 |
| | Proposed model vs. EMD-MOHHO-ELM | 49.820 | 46.416 | 52.879 | 3.568 | 46.575 | 34.448 | 7.018 |
| | Proposed model vs. CEEMD-MOHHO-ELM | 52.074 | 67.830 | 44.809 | 15.304 | 68.800 | 49.091 | 30.831 |
| Chongqing | Proposed model vs. ARIMA | 72.103 | 70.144 | 72.960 | 21.852 | 70.714 | 37.607 | 43.025 |
| | Proposed model vs. ELM | 72.699 | 71.733 | 74.923 | 26.864 | 71.724 | 40.747 | 55.096 |
| | Proposed model vs. LSSVM | 72.033 | 70.773 | 73.407 | 24.151 | 70.714 | 33.272 | 48.702 |
| | Proposed model vs. EMD-MOHHO-ELM | 67.149 | 64.848 | 70.025 | 12.929 | 64.035 | 49.164 | 24.712 |
| | Proposed model vs. CEEMD-MOHHO-ELM | 35.322 | 27.732 | 35.478 | 1.439 | 26.786 | 27.579 | 2.419 |

## 5 Discussions

With the aim of showing the predicative superiority and effectiveness of the developed hybrid model, DM (Diebold–Mariano) test and stability test are performed in this subsection 5.1. Additionally, in subsection 5.2 three multi-objective algorithms: MOPSO (multi-objective particle swarm optimization) [38], MOGOA (multi-objective grasshopper optimization algorithm) [39] and MSSA (multi-objective salp swarm algorithm) [40] and four common test functions (ZDT1, ZDT2, ZDT3, and ZDT1 with a linear front) [41-42] are all adopted to benchmark the performance of the developed MOHHO algorithm.

### 5.1. DM test and stability test

In this subsection, the hypothesis testing i.e. DM (Diebold-Mariano) test [43] is employed to show the superiority of the presented hybrid forecasting model compared with the five models. More detailed description of DM test can be seen in the literatures [44-46]. Moreover, a stability test: (VR) variance ratio [47] is also performed in this subsection in order to show the higher stable ability of forecasting models. The higher the variance ratio is, the higher the forecasting stability of the model is. **Table 6** lists the results of the DM test and VR. Moreover, the values marked in boldface represent the best values of every evaluation metric.

From the values in **Table 6**, several conclusions are obtained that the values of DM test are larger than the $Z_{0.01/2} = 2.58$, illustrating that the prediction ability of the developed hybrid method is different from that of the comparison models, meaning that our proposed hybrid forecasting model observably outperforms the comparison forecasting models. Additionally, the biggest VR values of the presented hybrid model displays that the proposed forecasting model possesses higher forecasting stability than the five models taken as comparisons. The discussions in this subsection once again prove that the developed hybrid forecasting model can enhance the prediction effectiveness, which can be used to reduce the atmospheric pollutants emissions and warn the public before the occurrence of hazardous air pollutants.

**Table 6**. Results of DM test and stability test of different models.

| Areas | Models | $PM_{2.5}$ | | | $PM_{10}$ | | |
|---|---|---|---|---|---|---|---|
| | | DM values | P values | VR | DM values | P values | VR |
| Jinan | ARIMA | 4.514 [a] | $4.003*10^{-5}$ | 0.571 | 4.396 [a] | $5.911*10^{-5}$ | 0.542 |
| | ELM | 4.413 [a] | $5.600*10^{-5}$ | 0.553 | 4.268 [a] | $8.992*10^{-5}$ | 0.422 |
| | LSSVM | 4.580 [a] | $3.207*10^{-5}$ | 0.277 | 6.416 [a] | $5.336*10^{-8}$ | 0.364 |
| | EMD-MOHHO-ELM | 5.118 [a] | $5.162*10^{-6}$ | 0.689 | 4.608 [a] | $2.919*10^{-5}$ | 0.609 |
| | CEEMD-MOHHO-ELM | 3.316 [a] | $1.726*10^{-3}$ | 0.600 | 3.966 [a] | $2.377*10^{-4}$ | 0.515 |
| | ICEEMDAN-MOHHO-ELM | — | — | **0.823** | — | — | **0.664** |
| Nanjing | ARIMA | 4.581 [a] | $3.202*10^{-5}$ | 0.556 | 4.686 [a] | $2.252*10^{-5}$ | 0.525 |
| | ELM | 3.839 [a] | $3.550*10^{-4}$ | 0.501 | 5.014 [a] | $7.380*10^{-6}$ | 0.625 |
| | LSSVM | 3.775 [a] | $4.324*10^{-4}$ | 0.458 | 4.880 [a] | $1.166*10^{-5}$ | 0.535 |
| | EMD-MOHHO-ELM | 4.945 [a] | $9.329*10^{-6}$ | 0.991 | 4.901 [a] | $1.084*10^{-5}$ | 0.872 |
| | CEEMD-MOHHO-ELM | 1.773 [c] | $8.245*10^{-2}$ | 0.594 | 1.815 [c] | $7.562*10^{-2}$ | 0.699 |
| | ICEEMDAN-MOHHO-ELM | | | **0.884** | | | **0.915** |
| Chongqing | ARIMA | 3.635 [a] | $4.003*10^{-4}$ | 0.790 | 3.863 [a] | $3.291*10^{-4}$ | 0.713 |
| | ELM | 3.987 [a] | $4.003*10^{-4}$ | 0.795 | 3.697 [a] | $5.503*10^{-4}$ | 0.663 |
| | LSSVM | 3.892 [a] | $4.003*10^{-4}$ | 0.746 | 3.747 [a] | $4.722*10^{-4}$ | 0.680 |
| | EMD-MOHHO-ELM | 4.381 [a] | $4.003*10^{-5}$ | 0.677 | 4.453 [a] | $4.898*10^{-5}$ | 0.772 |
| | CEEMD-MOHHO-ELM | 2.242 [b] | $4.003*10^{-2}$ | 0.843 | 3.410 [a] | $1.308*10^{-3}$ | 0.836 |
| | ICEEMDAN-MOHHO-ELM | — | — | **0.904** | — | — | **0.929** |

[a] is the 1% significance level $Z_{0.01/2} = 2.58$; [b] is the 5% significance level $Z_{0.05/2} = 1.96$

[c] is the 10% significance level $Z_{0.10/2} = 1.64$;

### 5.2. Validity of the proposed MOHHO algorithm

To verify the performance of the developed MOHHO algorithm, a set of four challenging multi-objective test functions, namely ZDT test problems (i.e. ZDT1, ZDT2, ZDT3 and ZDT1 with line front) as well as the well-known algorithm: MOPSO and the new proposed ones: MOGOA and MSSA are utilized to test the quality of the developed MOHHO algorithm. Additionally, the Inverted Generational Distance (IGD) [48] is taken as a performance metric in this subsection for quantitatively analysis the performance of the developed algorithm. The aforementioned algorithms are run 50 times and the statistical results are reported and shown in **Table 7** and **Figs 3-4**. It is worth noting that an archive size of 100, 100 iterations, and 40 search agents were set in each time. Moreover, the qualitative and quantitative results listed as below show that the proposed MOHHO performs well, which adds a new algorithm for multi-objective optimization problems.

$$\text{IGD} = \frac{1}{n}\sqrt{\sum_{t=1}^{n} d_t^2} \qquad (27)$$

Where $d_t$ is the Euclidean distance between the *t-th* true Pareto optimal solution and the nearest ones achieved by the MOPSO, MSSA, MOGOA and MOHHO, $n$ is the size of true Pareto optimal answers.

As seen from **Figs 3-4** and **Table 7**, several analyses can be obtained as below: (a) Inspecting **Figs 3-4**, it might be seen that MOGOA indicates the poorest convergence despite its good coverage in ZDT1 and ZDT1 with line front. Conversely, the MOHHO and MSSA can both converge well to all the true Pareto optimal fronts. It is worth mentioning that MOPSO algorithm significantly outperforms the other three algorithms on the majority of ZDT3 test function. (b) **Table 7** indicates that the MOHHO algorithm managed to outperform the MOPSO, MOGOA and MSSA algorithms significantly on the ZDT test functions, except for ZDT3 test function. The advantage can be observed in the columns of ZDT1, ZDT2 and ZDT1 with line front test functions, demonstrating a higher accuracy and better robustness of MOHHO compared to the other algorithms. As for ZDT3 test function, The MOHHO algorithm, however, indicates very competitive results in comparison with the listed algorithms and occasionally outperforms them.

**Remark.** The aforementioned qualitative and quantitative analyses and results show that MOHHO can be able to efficiently approximate the true front of ZDT test functions with a very high convergence and coverage. What's more, this algorithm adds a new way for addressing the multi-objective optimization issues, which can be regarded as an alternative for solving challenging real-world problems as well.

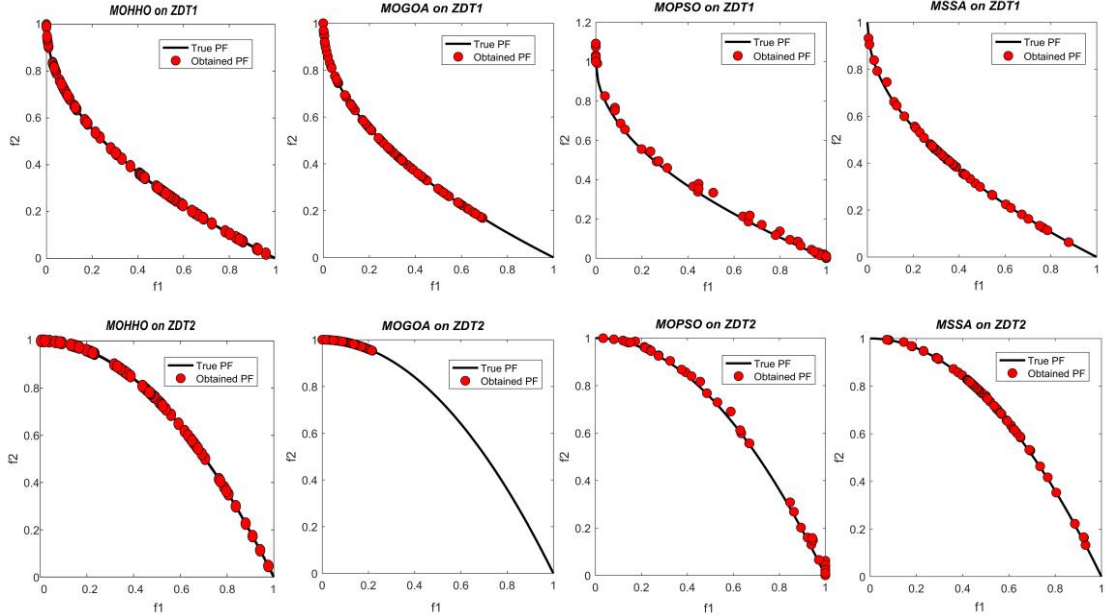

**Fig. 3.** The Pareto optimal solutions achieved by MOHHO, MOGOA, MOPSO and MSSA for ZDT1 and ZDT2. (The PF represents Pareto Font)

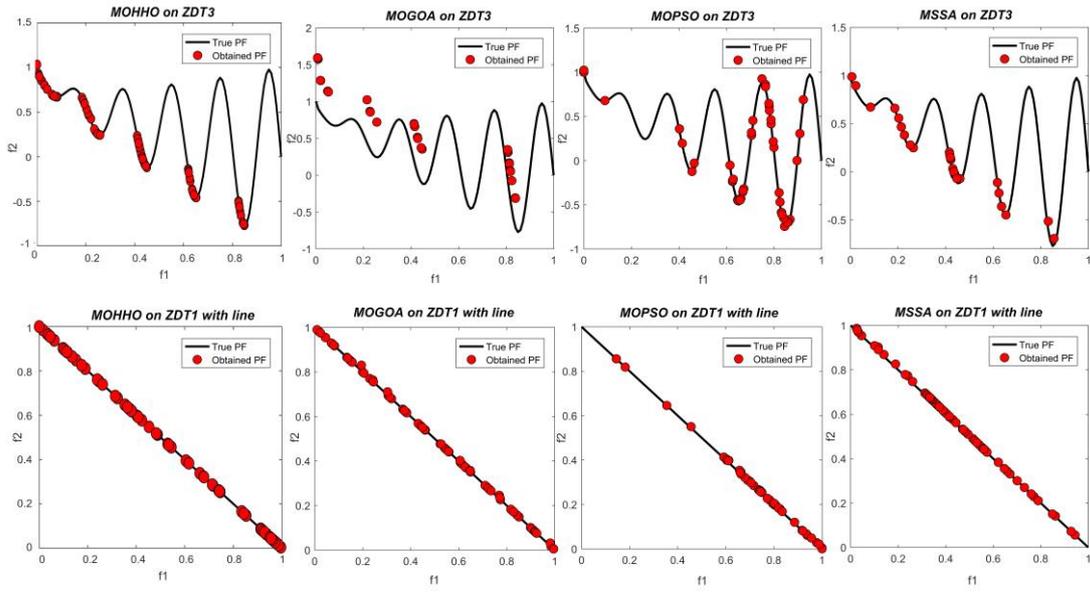

**Fig. 4.** The Pareto optimal solutions achieved by MOHHO, MOGOA, MOPSO and MSSA for ZDT3 and ZDT1 with linear front.

**Table 7**
The values of IGD of the multi-objective algorithms.

| Algorithm | ZDT1 | | | | | Algorithm | ZDT2 | | | | |
|---|---|---|---|---|---|---|---|---|---|---|---|
| | Ave. | Std. | Median | Best | Worst | | Ave. | Std. | Median | Best | Worst |
| MOGOA | 0.025581 | 0.016067 | 0.017007 | 0.011493 | 0.094338 | MOGOA | 0.058590 | 0.005903 | 0.057628 | 0.057628 | 0.098954 |
| MOPSO | 0.008471 | 0.014866 | 0.003294 | 0.001799 | 0.069642 | MOPSO | 0.022392 | 0.030090 | 0.003447 | 0.001649 | 0.080115 |
| MSSA | 0.002391 | 0.000848 | 0.002134 | 0.001603 | 0.005611 | MSSA | 0.002459 | 0.001142 | 0.002189 | 0.001716 | 0.009682 |
| MOHHO | **0.001525** | **0.000309** | **0.001441** | **0.001158** | **0.002594** | MOHHO | **0.001485** | **0.000209** | **0.001449** | **0.001155** | **0.002188** |

| Algorithm | ZDT3 | | | | | Algorithm | ZDT1 with linear front | | | | |
|---|---|---|---|---|---|---|---|---|---|---|---|
| | Ave. | Std. | Median | Best | Worst | | Ave. | Std. | Median | Best | Worst |
| MOGOA | 0.021405 | 0.000997 | 0.021442 | 0.019231 | 0.023190 | MOGOA | 0.006272 | 0.015994 | 0.001538 | 0.001238 | 0.094659 |
| MOPSO | **0.008227** | 0.002943 | **0.007529** | **0.005115** | **0.018103** | MOPSO | 0.020488 | 0.025154 | 0.006306 | 0.001857 | 0.092015 |
| MSSA | 0.024277 | 0.000581 | 0.024342 | 0.023057 | 0.025492 | MSSA | 0.002216 | 0.000445 | 0.002140 | 0.001522 | 0.003953 |
| MOHHO | 0.024593 | **0.000307** | 0.024496 | 0.024127 | 0.025513 | MOHHO | **0.001512** | **0.000311** | **0.001468** | **0.001076** | **0.002787** |

**Note:** The values with bold represent the best ones of the IGD.

# 6 Conclusions

Along with the accelerating of urbanization and industrialization process, and the increasing growth of energy consumption, air pollution issues are becoming more and more serious, which receive increasing attention in the world. High levels of air pollution can seriously affect people's living environment and even endanger their lives. Thus, it is urgent to design the accurate and reliable air pollutant forecasting model, which can be utilized to help reduce air pollution and guide people's daily activities and warn the public before the occurrence of hazardous air pollutants.

Recently, most forecasting air pollutant concentrations forecasting models have been developed and eventually enhanced the forecasting effectiveness to some extent. Nevertheless, many deficiencies, such as simple data decomposition techniques, ignoring the importance of predictive stability, and poor initial parameters of models and so on, still exist in the aforementioned models, which have significantly effect on the performance of air pollution prediction. Hence, to address these issues, a novel hybrid model based on the improvement of the prediction accuracy and stability of forecasting models is proposed in this study. Specifically, a powerful data preprocessing techniques is applied to decompose the original time series into different modes from low- frequency to high- frequency. Next, a new multi-objective algorithm called MOHHO is first developed in this study, which are designed to tune the parameters of the ELM model with the hope of archiving high accuracy and stability for air pollutant concentrations prediction at the same time. And the optimized ELM model is used to perform the time series prediction. Finally, a scientific and robust evaluation system including several error criteria, benchmark models, and several experiments using six air pollutant concentrations time series from three cities in China is designed to make a compressive evaluation for the proposed hybrid model to deeply confirm the superiority of the presented hybrid model in terms of the prediction ability.

As a result, according to the empirical results and the aforementioned discussions, conclusions can be drawn that the developed hybrid forecasting model possesses higher prediction accuracy and more stable results than the five models considered as comparisons. The hybrid model proposed in this study adds a novel feasible measure for air pollution prediction, and its superior prediction ability may help to develop effective plans for air pollutant emissions and prevent health problems caused by air pollution.

**Appendix A:** *Concepts of multi-objective optimization*

Owing to two or more objective functions in MOPs, the conventional relational operators, such as $>$, $\geq$, $<$, $\leq$ and $=$ etc., cannot be suitable for multi-objective optimization. Fortunately, a new concept of dominates is proposed by scientists, which can be adopted to search for best solutions. In the multi-objective optimization, a minimization problem can be expressed as below:

$$\begin{aligned} Minimize: \ & f(\boldsymbol{x}) = [f_1(\boldsymbol{x}), f_2(\boldsymbol{x}), ..., f_{ob}(\boldsymbol{x})] \\ S.t. \ & g_j(\boldsymbol{x}) \geq 0, \quad j = 1, 2, ..., m \\ & h_j(\boldsymbol{x}) = 0, \quad j = 1, 2, ..., p \\ & L_j \leq x_j \leq U_j, \quad j = 1, 2, ..., n \end{aligned} \quad (B1)$$

In the above equation, the *n*, *ob*, *m* and *p* stand for the number of the variables, the number of the objective functions, the number of the inequality constraints and the number of the equality constraints, respectively. Meanwhile, $g_j(\bullet)$ and $h_j(\bullet)$ are the *j-th* inequality and the *j-th* equality constraints, respectively. $L_j$ and $U_j$ are the lower and upper boundary values of the *j-th* variable, respectively.

Moreover, the other concepts involved in the multi-objective optimization problems, such as the Pareto Dominance, the Pareto optimality, the Pareto optimal set, and the Pareto optimal front can be expressed by the following definitions:

**Definition 1.** Pareto Dominance:

$\boldsymbol{x} = (x_1, x_2, ..., x_n)$ dominates $\boldsymbol{y} = (y_1, y_2, ..., y_n)$ i.e. $\boldsymbol{x} \succ \boldsymbol{y}$ if and only if:

$$\forall i \in [1,n], [f(x_i) \geq f(y_i)] \wedge [\exists i \in [1,n]: f(x_i)] \tag{B2}$$

**Definition 2.** Pareto optimality:

A Pareto-optimal $\boldsymbol{x} \in \boldsymbol{X}$ iff:

$$\nexists \boldsymbol{y} \in \boldsymbol{X} \ s.t. \ F(\boldsymbol{y}) \succ F(\boldsymbol{x}) \tag{B3}$$

**Definition 3.** Pareto optimal set:

$$P_s := \{\boldsymbol{x}, \boldsymbol{y} \in \boldsymbol{X} \mid \exists F(\boldsymbol{y}) \succ F(\boldsymbol{x})\} \tag{B4}$$

**Definition 4.** Pareto optimal front:

A set including the value of objective functions for Pareto solutions set:

$$P_f := \{F(\boldsymbol{x}) \mid \boldsymbol{x} \in P_s\} \tag{B5}$$

**Appendix B. List of abbreviations**

**Table B1** The improvement percentages of criteria.

| Metric | Definition | Equation |
|---|---|---|
| $P_{MAE}$ | The improvement percentages of **MAE** | $P_{MAE} = \left|\dfrac{MAE_1 - MAE_2}{MAE_1}\right|$ |
| $P_{RMSE}$ | The improvement percentages of **RMSE** | $P_{RMSE} = \left|\dfrac{RMSE_1 - RMSE_2}{RMSE_1}\right|$ |
| $P_{MAPE}$ | The improvement percentages of **MAPE** | $P_{MAPE} = \left|\dfrac{MAPE_1 - MAPE_2}{MAPE_1}\right|$ |
| $P_{IA}$ | The improvement percentages of **IA** | $P_{IA} = \left|\dfrac{IA_1 - IA_2}{IA_1}\right|$ |
| $P_{U1}$ | The improvement percentages of **U1** | $P_{U1} = \left|\dfrac{U1_1 - U1_2}{U1_1}\right|$ |
| $P_{U2}$ | The improvement percentages of **U2** | $P_{U2} = \left|\dfrac{U2_1 - U2_2}{U2_1}\right|$ |
| $P_r$ | The improvement percentages of ***r*** | $P_r = \left|\dfrac{r_1 - r_2}{r_1}\right|$ |

**Acknowledgements**

This work was supported by the National Social Science Foundation of China

(Grant No. 17ZDA093).

**References**

[1] Tian, G., Qiao, Z., Xu, X., 2014. Characteristics of particulate matter ($PM_{10}$) and its relationship with meteorological factors during 2001-2012 in Beijing. Environmental Pollution. doi:10.1016/j.envpol.2014.04.036

[2] Qin, S., Liu, F., Wang, J., Sun, B., 2014. Analysis and forecasting of the particulate matter (PM) concentration levels over four major cities of China using hybrid models. Atmospheric Environment. doi:10.1016/j.atmosenv.2014.09.046

[3] Song, Y., Qin, S., Qu, J., Liu, F., 2015. The forecasting research of early warning systems for atmospheric pollutants: A case in Yangtze River Delta region. Atmospheric Environment. doi:10.1016/j.atmosenv.2015.06.032

[4] Wang, Q., Wang, Y., Zhou, P., Wei, H., 2016. Whole process decomposition of energy-related $SO_2$ in Jiangsu Province, China. Applied Energy. doi:10.1016/j.apenergy.2016.05.073

[5] Niu, M., Wang, Y., Sun, S., Li, Y., 2016. A novel hybrid decomposition-and-ensemble model based on CEEMD and GWO for short-term $PM_{2.5}$ concentration forecasting. Atmospheric Environment. doi:10.1016/j.atmosenv.2016.03.056

[6] Zhang, H., Wang, Y., Hu, J., Ying, Q., Hu, X.M., 2015. Relationships between meteorological parameters and criteria air pollutants in three megacities in China. Environmental Research. doi:10.1016/j.envres.2015.04.004

[7] Chen, Z., Wang, J.-N., Ma, G.-X., Zhang, Y.-S., 2013. China tackles the health effects of air pollution. The Lancet. doi:10.1016/s0140-6736(13)62064-4

[8] Asian Development Bank. Toward an environmentally sustainable future: country environmental analysis of the People's Republic of China. Manila: Asian Development Bank; 2013

[9] Biancofiore, F., Busilacchio, M., Verdecchia, M., Tomassetti, B., Aruffo, E., Bianco, S., Di Tommaso, S., Colangeli, C., Rosatelli, G., Di Carlo, P., 2017. Recursive neural network model for analysis and forecast of PM10 and PM2.5. Atmospheric Pollution Research. doi:10.1016/j.apr.2016.12.014

[10] Xu, Y., Du, P., Wang, J., 2017. Research and application of a hybrid model based on dynamic fuzzy synthetic evaluation for establishing air quality forecasting and early warning system: A case study in China. Environmental Pollution. doi:10.1016/j.envpol.2017.01.043

[11] Zhang, L., Lin, J., Qiu, R., Hu, X., Zhang, H., Chen, Q., Tan, H., Lin, D., Wang, J., 2018. Trend analysis and forecast of $PM_{2.5}$ in Fuzhou, China using the ARIMA model. Ecological Indicators. doi:10.1016/j.ecolind.2018.08.032

[12] You, M.L., Shu, C.M., Chen, W.T., Shyu, M.L., 2017. Analysis of cardinal grey relational grade and grey entropy on achievement of air pollution reduction by evaluating air quality trend in Japan. Journal of Cleaner Production. doi:10.1016/j.jclepro.2016.10.072

[13] Ma, X., Xie, M., Wu, W., Zeng, B., Wang, Y., Wu, X., 2019. The novel fractional discrete multivariate grey system model and its applications. Applied Mathematical Modelling. doi:10.1016/j.apm.2019.01.039

[14] Ma, X., Mei, X., Wu, W., Wu, X., Zeng, B., 2019. A novel fractional time delayed grey model with Grey Wolf Optimizer and its applications in forecasting the natural gas and coal consumption in Chongqing China. Energy. doi:10.1016/j.energy.2019.04.096

[15] Bai, Y., Zeng, B., Li, C., Zhang, J., 2019. An ensemble long short-term memory neural network for hourly PM2.5 concentration forecasting. Chemosphere. doi:10.1016/j.chemosphere.2019.01.121

[16] Hao, Y., Tian, C., 2019. The study and application of a novel hybrid system for air quality early-warning. Applied Soft Computing Journal. doi:10.1016/j.asoc.2018.09.005

[17] Li, H., Wang, J., Li, R., Lu, H., 2019. Novel analysis–forecast system based on multi-objective optimization for air quality index. Journal of Cleaner Production. doi:10.1016/j.jclepro.2018.10.129

[18] Du, P., Wang, J., Guo, Z., Yang, W., 2017. Research and application of a novel hybrid forecasting system based on multi-objective optimization for wind speed forecasting. Energy Conversion and Management. doi:10.1016/j.enconman.2017.07.065

[19] Wang, J., Liu, F., Song, Y., Zhao, J., 2016. A novel model: Dynamic choice artificial neural network (DCANN) for an electricity price forecasting system. Applied Soft Computing Journal. doi:10.1016/j.asoc.2016.07.011

[20] Du, P., Wang, J., Yang, W., Niu, T., 2018. Multi-step ahead forecasting in electrical power system using a hybrid forecasting system. Renew. Energy. https://doi.org/10.1016/j.renene.2018.01.113

[21] Wang, D., Wei, S., Luo, H., Yue, C., Grunder, O., 2017. A novel hybrid model for air quality index forecasting based on two-phase decomposition technique and modified extreme learning machine. Sci. Total Environ. https://doi.org/10.1016/j.scitotenv.2016.12.018

[22] Zhou, Q., Jiang, H., Wang, J., Zhou, J., 2014. A hybrid model for PM$_{2.5}$ forecasting based on ensemble empirical mode decomposition and a general regression neural network. Sci. Total Environ. https://doi.org/10.1016/j.scitotenv.2014.07.051

[23] Colominas, M.A., Schlotthauer, G., Torres, M.E., 2014. Improved complete ensemble EMD: A suitable tool for biomedical signal processing. Biomedical Signal Processing and Control. doi:10.1016/j.bspc.2014.06.009

[24] Xu, Y., Yang, W., Wang, J., 2017. Air quality early-warning system for cities in China. Atmospheric Environment. doi:10.1016/j.atmosenv.2016.10.046

[25] Huang, N.E., Shen, Z., Long, S.R., Wu, M.C., Snin, H.H., Zheng, Q., Yen, N.C., Tung, C.C., Liu, H.H., 1998. The empirical mode decomposition and the Hubert spectrum for nonlinear and non-stationary time series analysis. Proceedings of the Royal Society A: Mathematical, Physical and Engineering Sciences. doi:10.1098/rspa.1998.0193

[26] Wu, Z., Huang, N.E., 2008. Ensemble empirical mode decomposition: a noiseassisted data analysis method. Adv. Adapt. Data Anal. 1, 1e41.

[27] Yeh, J.R., Shieh, J.S., Huang, N.E., 2010. Complementary ensemble empirical mode decomposition: a novel noise enhanced data analysis method. Adv. Adapt. Data Anal. 2, 135e156.

[28] Torres, M.E., Colominas, M.A., Schlotthauer, G., Flandrin, P., 2011. A complete ensemble empirical mode decomposition with adaptive noise, in: ICASSP, IEEE International Conference on Acoustics, Speech and Signal Processing - Proceedings. doi:10.1109/ICASSP.2011.5947265

[29] Wang, J., Yang, W., Du, P., Li, Y., 2018. Research and application of a hybrid forecasting framework based on multi-objective optimization for electrical power system. Energy. doi:10.1016/j.energy.2018.01.112

[30] Li, X., Li, C., 2016. Improved CEEMDAN and PSO-SVR Modeling for Near-Infrared Noninvasive Glucose Detection. Computational and Mathematical Methods in Medicine. doi:10.1155/2016/8301962

[31] Du, P., Wang, J., Yang, W., Niu, T., 2019. A novel hybrid model for short-term wind power forecasting. Applied Soft Computing Journal. doi: 10.1016/j.asoc.2019.03.035

[32] Zhang, J., Guo, Y., Shen, Y., Zhao, D., Li, M., 2018. Improved CEEMDAN-wavelet transform de-noising method and its application in well logging noise reduction. Journal of Geophysics and Engineering. doi:10.1088/1742-2140/aaa076

[33] Heidari, A.A., Mirjalili, S., Faris, H., Aljarah, I., Mafarja, M., Chen, H., 2019. Harris hawks optimization: Algorithm and applications. Future Generation Computer Systems. doi:10.1016/j.future.2019.02.028

[34] Huang G-B, Zhu Q, Siew C, Ã GH, Zhu Q, Siew C, et al. Extreme learning machine: Theory and applications. Neurocomputing 2006; 70:489–501. doi: 10.1016/j.neucom.2005.12.126.

[35] Peng T, Zhou J, Zhang C, Zheng Y. Multi-step ahead wind speed forecasting using a hybrid model based on two-stage decomposition technique and AdaBoost-extreme learning machine. Energy Convers Manag 2017; 153:589–602. doi: 10.1016/j.enconman.2017.10.021.

[36] Li S, Goel L, Wang P. An ensemble approach for short-term load forecasting by extreme learning machine. Appl Energy 2016; 170:22–9. doi: 10.1016/j.apenergy.2016.02.114.

[37] Yang, W., Wang, J., Lu, H., Niu, T., Du, P., 2019. Hybrid wind energy forecasting and analysis system based on divide and conquer scheme: A case study in China. Journal of Cleaner Production. doi:10.1016/j.jclepro.2019.03.036

[38] Coello Coello, C.A., Lechuga, M.S., 2002. MOPSO: A proposal for multiple objective particle swarm optimization, in: Proceedings of the 2002 Congress on Evolutionary Computation, CEC 2002. doi:10.1109/CEC.2002.1004388

[39] Mirjalili, S.Z., Mirjalili, S., Saremi, S., Faris, H., Aljarah, I., 2018. Grasshopper optimization algorithm for multi-objective optimization problems. Applied Intelligence. doi:10.1007/s10489-017-1019-8

[40] Mirjalili, S., Gandomi, A.H., Mirjalili, S.Z., Saremi, S., Faris, H., Mirjalili, S.M., 2017. Salp Swarm Algorithm: A bio-inspired optimizer for engineering design problems. Advances in Engineering Software. doi:10.1016/j.advengsoft.2017.07.002

[41] Tian, C., Hao, Y., Hu, J., 2018. A novel wind speed forecasting system based on hybrid data preprocessing and multi-objective optimization. Applied Energy. doi:10.1016/j.apenergy.2018.09.012

[42] Yang, W., Wang, J., Niu, T., Du, P., 2019. A hybrid forecasting system based on a dual decomposition strategy and multi-objective optimization for electricity price forecasting. Applied Energy. doi:10.1016/j.apenergy.2018.11.034

[43] Diebold, F.X., Mariano, R.S., 1995. Comparing predictive accuracy. Journal of Business and Economic Statistics. doi:10.1080/07350015.1995.10524599

[44] Wang, J., Du, P., Lu, H., Yang, W., Niu, T., 2018. An improved grey model optimized by multi-objective ant lion optimization algorithm for annual electricity consumption forecasting. Applied Soft Computing Journal. doi:10.1016/j.asoc.2018.07.022

[45] Sun, S., Wei, Y., Tsui, K.L., Wang, S., 2019. Forecasting tourist arrivals with machine learning and internet search index. Tourism Management. doi:10.1016/j.tourman.2018.07.010

[46] Niu, X., Wang, J., 2019. A combined model based on data preprocessing strategy and multi-objective optimization algorithm for short-term wind speed forecasting. Applied Energy. doi:10.1016/j.apenergy.2019.03.097

[47] Yang, Z., Wang, J., 2017. A new air quality monitoring and early warning system: Air quality assessment and air pollutant concentration prediction. Environmental Research. doi:10.1016/j.envres.2017.06.002

[48] Mirjalili, S., Jangir, P., Saremi, S., 2017. Multi-objective ant lion optimizer: a multi-objective optimization algorithm for solving engineering problems. Applied Intelligence. doi:10.1007/s10489-016-0825-8